\documentclass[fleqn,10pt]{wlscirep}
\usepackage[utf8]{inputenc}
\usepackage[T1]{fontenc}
\usepackage{enumitem}
\usepackage{multirow}
\usepackage[normalem]{ulem}

\title{Convolutional Spiking Neural Networks for Intent Detection Based on 
Anticipatory Brain Potentials Using Electroencephalogram}

\author[1]{Nathan Lutes}
\author[2]{Venkata Sriram Siddhardh Nadendla}
\author[1,*]{K. Krishnamurthy}
\affil[1]{Missouri University of Science and Technology, Mechanical and Aerospace Engineering, Rolla, MO 65409, U.S.A.}
\affil[2]{Missouri University of Science and Technology, Computer Science, Rolla, MO 65409, U.S.A.}

\affil[*]{kkrishna@mst.edu}

\begin{document}

\begin{abstract}
Spiking neural networks (SNNs) are receiving increased attention because they mimic synaptic connections in biological systems and produce spike trains, which can be approximated by binary values for computational efficiency. Recently, the addition of convolutional layers to combine the feature extraction power of convolutional networks with the computational efficiency of SNNs has been introduced. This paper studies the feasibility of using a convolutional spiking neural network (CSNN) to detect anticipatory slow cortical potentials (SCPs) related to braking intention in human participants using an electroencephalogram (EEG). 
Data was collected during an experiment wherein participants operated a remote-controlled vehicle on a testbed designed to simulate an urban environment. Participants were alerted to an incoming braking event via an audio countdown to elicit anticipatory potentials that were measured using an EEG. The CSNN’s performance was compared to a standard CNN, EEGNet and three graph neural networks via 10-fold cross-validation. The CSNN outperformed all the other neural networks, and had a predictive accuracy of 99.06\% with a true positive rate of 98.50\%, a true negative rate of 99.20\% and an F1-score of 0.98. Performance of the CSNN was comparable to the CNN in an ablation study using a subset of EEG channels that localized SCPs. Classification performance of the CSNN degraded only slightly when the floating-point EEG data were converted into spike trains via delta modulation to mimic synaptic connections. 
\end{abstract}

\flushbottom
\maketitle
%
%
\thispagestyle{empty}

\section*{Introduction}
Significant advancements in computing hardware, such as graphics processing units and field-programmable gate arrays, along with the availability of large datasets, has enabled researchers to develop highly effective neural networks in the last decade. However, training and utilizing these networks often involves a large amount of energy consumption, thus restricting the deployment of neural networks for data, and/or energy, limited settings: typically applications in dynamic/mobile environments. On the contrary, biology-inspired neural networks need only very few or even only one data point to perform at a competitive level compared to "traditional" neural networks (see page 54 in ref \cite{Christensen_2022}). Therefore, machine learning architectures more closely resembling biological neural networks are quickly gaining in popularity. One such example is the spiking neural network (SNN) \cite{tan_sarlija_kasabov_2020, dora_kasabov_2021, tavanaei_ghodrati_kheradpisheh_masquelier_maida_2019} which mimics biological neural networks through its layers composed of \emph{spiking neurons}. These neurons more closely resemble the synaptic connections between neurons in biological neural networks through their emission of aperiodic spikes as opposed to floating point numbers in the case of the traditional artificial neuron. This sparse and discrete behavior of SNNs has been shown to reduce energy consumption by orders of magnitude when implemented on emerging neuromorphic hardware \cite{davies_wild_orchard_sandamirskaya_guerra_joshi_plank_risbud_2021}. However, shallow SNNs can be insufficient to detect patterns that occur at random times/locations in tasks such as object detection/segmentation, similar to standard multi-layer perceptrons. This has inspired the development of hybrid convolutional and spiking neural networks, referred to as convolutional spiking neural networks (CSNNs) \cite{8354825, vaila_chiasson_saxena_2019, lee_panda_srinivasan_roy_2018}, which combine the convolutional layer's power of extracting spatio-temporal features with the energy efficiency of spiking neuron layers. In the past few years, CSNNs have received increased attention in diverse applications such as computer vision \cite{barchid_mennesson_djeraba_2021, 1198140}, speech recognition \cite{dong_huang_xu_2018}, hand-gesture recognition \cite{xing_di_caterina_soraghan_2020} and detection of Alzheimer's disease \cite{turkson_qu_mawuli_eghan_2021} reinforcing their utility in deciphering complex and multi-dimensional data.

The main contribution of this study is to evaluate the use of CSNNs in \emph{advanced driver-assist systems} (ADAS), specifically those approaches that utilize electroencephalograms (EEGs). ADAS can be summarized as a group of assistive technologies designed to decrease the cognitive load associated with the driving task by assisting with driving and/or parking decisions thus aiding the driver in safely operating their vehicle. This technology has been rapidly introduced in modern vehicles and has shown to greatly improve road safety and reduce traffic accidents \cite{8620826}. EEG is a method of measuring and recording electrical potentials from across various points in the human brain, thus serving as a primary method of discerning a person's current cognitive activity. EEG-based applications are commonly explored in the field of brain-computer-interface (BCI) \cite{rashid_sulaiman_p_p_p_abdul_majeed_musa_ab_nasir_bari_khatun_2020}, which has contributed in the development of machine learning models dedicated specifically to the analysis and interpretation of EEG signals, for example "EEGNet" \cite{lawhern_2018}. The inclusion of EEG as an auxiliary input source effectively fuses the fields of ADAS and BCI and gives subsequently developed technologies the advantage of an accurate real-time measure of otherwise unknown aspects of the driver state\cite{chuang_lai_ko_kuo_lin_2010, lin_ko_chen_chen_lin_2010, zheng_lu_2017, hajinoroozi_zhang_huang_2017} and also allows for the prediction of a  driver's intended action (e.g. braking) \cite{10.3389/fneng.2012.00013, garipelli_chavarriaga_del_r_millan_2013,10.3389/fnins.2014.00222} before it occurs. Literature has reported anticipatory potentials being observed as early as 130 ms \cite{haufe_treder_gugler_sagebaum_curio_blankertz_2011} and 320 $\pm$ 200 ms \cite{khaliliardali_chavarriaga_gheorghe_millan_2015} before action onset. The present study focuses on the latter advantage of EEG and seeks to train a CSNN as the predictive classifier to detect these anticipatory brain potentials and thus predict braking intention. 

Although some initial studies have been made to demonstrate the effectiveness of shallow SNNs in typical BCI applications \cite{9669864, 9629621, 9024211, singanamalla_lin_2021, kumarasinghe_kasabov_taylor_2021, yan_zhou_wong_2022, shah_wang_doborjeh_doborjeh_kasabov_2019}, the proliferation of other convolutional networks in the realm of BCI (e.g. EEGNet) and the reported success of deep learning methods in EEG decoding problems \cite{9721187} implies that the inclusion of deep learning methods, such as the addition of convolutional layers, leads to a performance gain in classification tasks involving EEG data. Furthermore, the relative ease with which SNNs and their deep learning counterparts, CSNNs, can be mapped to emerging high-efficiency neuromorphic-computing hardware \cite{davies_wild_orchard_sandamirskaya_guerra_joshi_plank_risbud_2021, merolla_arthur_alvarez_icaza_cassidy_sawada_akopyan_jackson_imam_guo_nakamura_etal_2014, jo_chang_ebong_bhadviya_mazumder_lu_2010,  DBLP:journals/corr/abs-1906-08853, ivanov_2022} makes them ideally suited for deployment in mobile, energy-limited applications. The use of energy-efficient neuromorphic hardware becomes even more advantageous when implementing various learning methods for online continuous learning or one-shot learning\cite{scherr_stock1_maass_2020} in energy-constrained applications. 

To the author's knowledge, the potential of CSNNs for EEG-based ADAS has not yet been explored and will be a novel contribution. To achieve a fairer juxtaposition than directly comparing the CSNN's performance in this study to other methods in the literature, additional neural network models were trained on the same dataset to provide clearer context. These models include: i) a CNN of similar architecture; ii) EEGNet; and iii) three graph neural networks (GNNs). The CNN was chosen to be a direct comparison of the spiking architecture to a non-spiking architecture, the EEGNet was chosen as the "state of the art" benchmark model because of its previous history of generalizing better across different BCI paradigms and high performance achievement as compared to existing CNNs and traditional approaches \cite{lawhern_2018}. Lastly, the inclusion of GNNs was motivated as an alternative to standard CNN networks because of their similar performance on adjacent EEG decoding tasks \cite{demir_koike-akino_wang_haruna_erdogmus_2021, https://doi.org/10.48550/arxiv.2007.13484, https://doi.org/10.48550/arxiv.2006.08924, https://doi.org/10.48550/arxiv.1907.07835}.

{\bf Related Work.} Previous studies on BCI-based driver intent detection present a gamut of technical approaches that mainly differ in the pre-processing strategies and various classifiers used. A popular family of classifiers rely on a powerful technique called "discriminant analysis", where a predictive function of a certain family (linear, quadratic, etc.) is created using independent variables and regression coefficients and used to predict a dependent variable. For instance, Teng \emph{et al} \cite{teng_bi_liu_2018} used the sequential forward-floating search method to define a feature set from powers of frequency points across 16 EEG channels. These features were then used as input to a regularized linear discriminant analysis (RLDA) classifier to determine braking intention from normal driving with a reported accuracy over 94\%. In another study, a modality combination consisting of EEG, tibalis anterior electromyography (EMG) and brake pedal signal were used as input to a RLDA classifier for braking intent detection \cite{kim_kim_haufe_lee_2014}. Khaliliardali, \emph{et al} \cite{khaliliardali_chavarriaga_gheorghe_millan_2015,
https://www.biorxiv.org/content/10.1101/443390v1}
used a low frequency bandpass filter ranging from 0.1 Hz to 1 Hz and a quadratic discriminant analysis (QDA) classifier to classify braking intention. Haufe \emph{et al.} \cite{haufe_treder_gugler_sagebaum_curio_blankertz_2011} compared EEG, EMG and brake pedal response to determine the input feature that predicts braking intention the fastest, again using RLDA as the classifier. 

As a competitor to the "discriminant analysis" methods, the other popular classification methods in the literature use neural networks and deep neural networks. For example, Hernandez, \emph{et al} \cite{hernandez_mozos_ferrandez_antelis_2018} conducted a braking intention study using support vector machines and convolutional neural networks (CNNs) to differentiate normal driving and braking intention EEG signals achieving a reported average accuracy of 71\% and 72\% for support vector machines and CNNs, respectively. Nguyen, \emph{et al}  \cite{nguyen_chung_2019} compared EEG band power-based and autoregressive-based feature selection methods for braking intent detection using EEG signals as input to a multilayer perceptron neural network, reporting a better accuracy of 91\% with the autoregressive based method. Lee, \emph{et al} \cite{lee_kim_lee_2017} used recurrent convolutional neural networks (RCNNs) to predict braking intention from EEG data, achieving an AUC score of 0.86. It is evident from the literature that a variety of methods have been used with mixed results. Although there are some examples of neural network usage, the use of SNNs for the braking intention EEG decoding problem is noticeably absent.

The EEG pattern studied here is the contingent negative variation (CNV), which is a type of slow cortical potential (SCP) that occurs prior to movement in the central region of the brain. The CNV, in particular, manifests when a subject is given a warning stimulus followed closely by an imperial stimulus, or stimulus requiring an action. It is featured in previous movement intention literature that also focused on driver braking intent detection\cite{10.3389/fneng.2012.00013,khaliliardali_chavarriaga_gheorghe_millan_2015,
https://www.biorxiv.org/content/10.1101/443390v1}. However, the CNV is not the only EEG pattern used for intention detection in the literature. Event-related desynchronization (ERD) is an EEG phenomenon occurring in the mu and beta frequency bands up to two seconds before movement is realized. It is marked by a decrease in the spectral power of EEG within those bands that is not restored until after the movement is completed. Planelles \emph{et al}\cite{s141018172} conducted a study to find a suitable classifier for ERD stemming from a self-determined reaching movement in healthy patients reporting 72\% accuracy using an SVM classifer. Chamanzar \emph{et al}\cite{7930496} developed a novel algorithm for using ERDs to detect hand movement intention using adaptive wavelet transform. They reported a one second detection delay, a sensitivity of 88\% and a selectivity of 78\%. ERDs are also used for motor imagery decoding problems. Song \emph{et al}\cite{8616177} used a two-phase classifier design to reduce false positives in a motor imagery for rehabilitation application using ERDs as the EEG input. The reported results included a sensitivity of 61\% and a selectivity of 78\%. Indeed, ERD is a popular and useful EEG related signal that could serve as an alternative to slow cortical potentials for movement intention related applications, and the potential of CSNNs for ERD, or CNV in general, phenomenon has not yet been explored.

The use of Bereitschaftpotentials (readiness potentials) has also received attention within the field of EEG related BCI. This is a slowly building neural signal occurring 1-2 seconds before movement onset\cite{Nguyen16397}. To better understand how the readiness potential is connected with areas of the brain responsible for the motor preparation process, Nguyen \emph{et al}\cite{Nguyen16397} conducted a study simultaneously integrating acquired EEG and fMRI through computational modeling and determined that reciprocal connections between the SMA and anterior mid-cingulate cortex (aMCC) are important to maintain the sustained activity of the readiness potential before movement. Other works have used the readiness potential for movement intent detection. Mirzabagherian \emph{et al}\cite{MIRZABAGHERIAN2023107159} developed two convolutional neural networks composed of temporal-spatial, separable and depth-wise layers and used these networks to detect movement-related cortical potentials (MRCPs) indicating five different hand movements performed by patients with cervical spinal cord injury. They reported a classification accuracy of 71\% and 65\% for the Temporal-Spatial Convolutional Iterative Residual Network and Temporal-Spatial Convolutional Residual Network for the EEG\_All dataset and accuracies of 58\% and 68\% for the EEG\_Low frequency dataset. Gatti \emph{et al}\cite{Gatti492660} also studied harnessing MRCPs for movement speed and force intent detection. A four class dataset was created having subjects conduct a right hand palmar grasp task using 20\% and 60\% of the maximum voluntary contraction and either a 3 second slow grasp or 0.5 second fast grasp for each amount of force. Convolutional neural nets were used as the classification model achieving an overall accuracy of 84\%. Mussini and Di Russo\cite{PPR:PPR485383} investigated how anxiety can affect anticipatory brain functions by observing its effect on pre-stimulus ERP and the Bereitschaftspotential when performing tasks with and without feedback. Their results showed that high anxiety can diminish the presence of these signals but the addition of feedback can restore them to low anxiety levels.

\section*{Results \label{sec: Results}}
\subsection*{Identification of Braking Intention Signature in EEG Signals}
\begin{figure*}[!t]
\centering
\includegraphics[width = \textwidth]{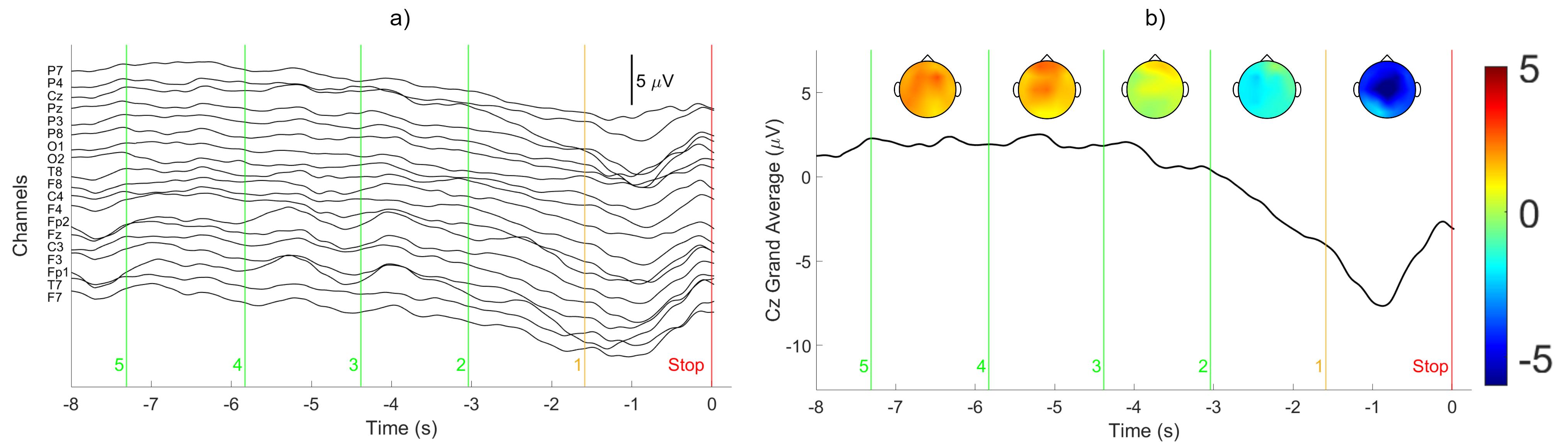}
\caption{Pre-Processed EEG Signals. (a) Channel potentials with associated countdown and "Stop" command markers and scale. (b) Cz grand average signal with scalp maps representing the grand average at the midpoint between two neighboring markers and color bar on the right displaying the potential in $\mu V$.}
\label{eeg_signal}
\end{figure*}

An example of the pre-processed EEG signals from 19 channels with data markers signifying the temporal locations of the audible countdown commands is shown in Figure \ref{eeg_signal}(a). Each marker is denoted by its associated countdown number from 5 to 1 and ending with "STOP" when the stop command was given. The Cz grand average of the pre-processed data is shown in Figure \ref{eeg_signal}(b) along with scalp plots visualizing how the channel grand averages changed over time. The grand averages were calculated by averaging the Cz electrode signal from all participants and across all trials, similar to the procedure followed by Khaliliardali, Chavarriaga, Gheorghe and Millan\cite{khaliliardali_chavarriaga_gheorghe_millan_2015}. The following observations can be made from Figure \ref{eeg_signal}.
\begin{enumerate}[leftmargin=3ex]
\setlength{\itemsep}{0.5ex}
    \item The negative EEG potential, termed contingent negative variation (CNV) potential, started after the "2" count marker and reached the maximum negative value between the "1" count marker and "Stop" command.
    \item The negativity rate sharply increased at the "1" count marker and the potential rate became sharply positive midway between the "1" count marker and "Stop" command. 
    \item Anticipatory potentials were clearly observed before the actual braking action.
    \item The more negative potentials were spatially localized in the centro-medial electrodes.
\end{enumerate}
The results obtained are consistent with other past studies on CNV \cite{10.3389/fneng.2012.00013,khaliliardali_chavarriaga_gheorghe_millan_2015,
https://www.biorxiv.org/content/10.1101/443390v1}.

\begin{table*}[!t]
\centering
\caption{Classification performance with floating-point EEG input data (best performance in each classification measure highlighted in bold font)}
\resizebox{\textwidth}{!}{
\begin{tabular}{c c c c c c c c c c c c}
\hline 
\\[-1.75ex]
\multirow{2}{*}{Model} & \multicolumn{2}{c}{Accuracy (\%)} & \multicolumn{2}{c}{TPR (\%)} & \multicolumn{2}{c}{TNR (\%)} & \multicolumn{2}{c}{F1-score} & Epochs & Train Time (s) & Inference
\\[0.5ex]
& Mean (SD) & \emph{p}-value & Mean (SD) & \emph{p}-value  & Mean (SD) & \emph{p}-value & Mean (SD) & \emph{p}-value & Mean (SD) & Mean (SD) & Time (s)
\\[0.75ex]
\hline\hline
\\[-1.5ex]
CSNN & $\textbf{99.06}$ (0.32) & $N/A^{\dagger}$ & $\textbf{98.50}$ (1.05) & $N/A^{\dagger}$ & $\textbf{99.20}$ (0.28) & $N/A^{\dagger}$ & $\textbf{0.98}$ (0.01) & $N/A^{\dagger}$ & 270 (87) & 17756 (5680) & 2.533
\\[1ex]
CNN & 79.46 (30.67) & 0.088 & 79.29 (39.66) & 0.18 & 79.50 (39.75) & 0.172 & 0.61 (0.41) & (0.043) & 130 (63) & 2348 (13) & 0.002
\\[1ex]
EEGNet & 97.64 (0.36) & $<$ 0.001 & 93.61 (2.03) & $<$ 0.001 & 98.64 (0.38) & 0.01 & 0.94 (0.01) & $<$ 0.001 & 50 (0) & 7238 (0.1) & 0.767
\\[1ex]
GCS & 57.09 (28.31)& 0.002 & 36.93 (46.00) & 0.003 & 62.10 (46.75) & 0.041 & 0.13 (0.16) & $<$ 0.001 & 110 (10) & 350 (34) & 0.293
\\[1ex]
GCN & 62.03 (27.51) & 0.003 & 44.83 (45.74) & 0.007 & 66.31 (43.79) & 0.051 & 0.21 (0.23) & $<$ 0.001 & 96 (15) & 281 (45) & 0.024
\\[1ex]
GIN & 54.66 (17.08) & $<$ 0.001 & 45.97 (29.57) & $<$ 0.001 & 56.82 (28.56) & 0.002 & 0.25 (0.10) & $<$ 0.001 & 82 (29) & 926 (328) & 0.759
\\[1ex]
\hline
\\[-2ex]
$^\dagger$ \text{Proposed model}
\end{tabular}
}
\label{StandardResults}
\end{table*}

\begin{table*}[!t]
\centering
\caption{Five-channel ablation study with floating-point EEG input data (best performance in each classification measure highlighted in bold font)}
\resizebox{\textwidth}{!} 
{
\begin{tabular}{c c c c c c c c c c c c}
\hline 
\\[-1.75ex]
\multirow{2}{*}{Model} & \multicolumn{2}{c}{Accuracy (\%)} & \multicolumn{2}{c}{TPR (\%)} & \multicolumn{2}{c}{TNR (\%)} & \multicolumn{2}{c}{F1-score} & Epochs & Train Time (s) & Inference
\\[0.5ex]
& Mean (SD) & \emph{p}-value & Mean (SD) & \emph{p}-value  & Mean (SD) & \emph{p}-value & Mean (SD) & \emph{p}-value & Mean (SD) & Mean (SD) & Time (s)
\\[0.75ex]
\hline\hline
\\[-1.5ex]
CSNN & 99.07 (0.27) & $N/A^{\dagger}$ & 98.35 (0.90) & $N/A^{\dagger}$ & $\textbf{99.24}$ (0.22) & $N/A^{\dagger}$ & $\textbf{0.98}$ (0.01) & $N/A^{\dagger}$ & 338 (77) & 20772 (4692) & 2.494
\\[1ex]
CNN & $\textbf{99.11}$ (0.32) & 0.618 & $\textbf{98.97}$ (0.91) & 0.128 & 99.15 (0.29) & 0.270 & 0.98 (0.01) & 0.582 & 999 (0) & 2365 (12) & 0.002
\\[1ex]
EEGNet & 92.22 (0.64) & $<$ 0.001 & 64.00 (4.10) & $<$ 0.001 & 99.22 (0.36) & 0.856 & 0.77 (0.03) & $<$ 0.001 & 50 (0) & 7239 (0.1) & 0.121
\\[1ex]
GCS & 65.67 (25.34) & 0.003 & 45.0 (41.53) & 0.004 & 71.07 (39.87) & 0.063 & 0.29 (0.26) & $<$ 0.001 & 118 (36) & 7 (1) & 0.144
\\[1ex]
GCN & 61.78 (25.57) & 0.001 & 45.0 (43.78) & 0.004 & 62.5 (35.84) & 0.01 & 0.29 (0.30) & $<$ 0.001 & 165 (99) & 7 (2) & 0.024
\\[1ex]
GIN & 52.67 (26.05) & $<$ 0.001 & 60 (45.95) & 0.027 & 50.89 (41.64) & 0.005 & 0.28 (0.22) & $<$ 0.001 & 94 (38.56) & 47 (10) & 0.641
\\[1ex]
\hline
\\[-2ex]
$^\dagger$ \text{Proposed model}
\end{tabular}
}
\label{AblationResults}
\end{table*}

\subsection*{Classification Performance - Case 1: 32-bit Single Precision Floating-Point EEG Input Data}

The final pre-processed dataset used for training the models included 10702 data segments collected from 15 participants in an experiment described in the Experimental Design section and preprocessed according to the procedure outlined in the Data Preprocessing section: 8573 data segments labelled as class `0' (no intention signal) and 2129 data segments labelled as class `1' (intention signal). The models were evaluated by means of 10-fold stratified cross-validation where the training and testing partitions of each fold maintained the original class distribution. The data segments in each fold were shuffled before training and testing. To mitigate the skewed distribution of the classes, wherein approximately 80\% of the dataset was negative class and approximately 20\% was positive class, the samples belonging to each class were given a weight used in the training loss function. The class weights were calculated as:

\noindent
\begin{equation}
w_{c,i} = \frac{N_{D}}{2N_{c,i}}
\end{equation}
where $N_{D}$ and $N_{c,i}$ are the total number of data points in the training partition and the number of class $i$ data points in the training partition, respectively. 

The convolutional layers in the CSNN and CNN contained a kernel size of $5\times5$, a stride of 1, a padding of 0, and 12 and 64 filters for the first and second convolutional layers, respectively. All max pooling layers of both architectures used a $2\times2$ pooling region with a stride of 2.
The sigmoid surrogate function smoothness value, $k$, for the surrogate backpropagation of the CSNN was chosen as $0.25$. 
The GCN and GCS graph convolutional layers had output sizes of 115, 28, 14 and 3. The GIN had graph convolutional layers with output sizes of 924, 462 and 231. The multi-layer perceptrons used in the GIN contained 5 hidden layers, each with 256 hidden neurons. All GNN architectures utilized in this work were implemented using the Spektral Python package \cite{grattarola_alippi_2021}. The CSNN and CNN models were implemented using PyTorch \cite{NEURIPS2019_9015}. EEGNet was implemented using tensorflow \cite{tensorflow2015-whitepaper} with default values with the exception of the length of the 2D convolutional kernel, which was set to 250.

All architectures were trained for up to 1000 epochs per fold with a batch size of 8, and with early termination if the loss did not improve for 50 subsequent epochs. The architectures had the same respective weight initialization for each fold. The CSNN Leaky-Integrate-and-Fire (LIF) layer had a memory decay rate of 0.5, a spiking threshold of 0.5 and used 25 input steps.
All models used the 'ADAM' optimizer with learning rate $\gamma = 5e-4$, running average coefficients $\beta_{1} = 0.9$ and $\beta_{2} = 0.999$, stability parameter $\epsilon = 1e-8$, and binary or categorical (for EEGNet) cross entropy as the loss function. 

Table \ref{StandardResults} shows the mean and standard deviation for predictive accuracy (Acc), true positive rate (TPR), true negative rate (TNR), F1-score, number of epochs trained, and total training time for each model. The table also shows each model's inference time for a single data point. The \emph{p}-values from two-sample t-tests comparing the CSNN with the other models are shown for each classification metric. No attempt was made to optimize the learning parameters of each network. The results indicate that the CSNN outperformed all the other models in every classification metric category albeit closely followed by EEGNet. The CSNN also showed small standard deviations showcasing the consistency in results across folds. The GNNs exhibited a tendency to get stuck in local minima, evidenced by large standard deviations in all the performance metrics. However, the CSNN had the largest average training time and largest inference time out of all of the models.

\subsection*
{Classification Performance - Case 2: Ablation Study with 32-bit Single Precision Floating-Point EEG Input Data from Five Channels}

Laboratory experiments offer the privilege of using state-of-the-art equipment that is typically unhindered in its data collection methods. However, real-world scenarios may present unique challenges where a full 20 channel EEG headset is not possible or practical. To study how a reduced number of available channels would affect the performance of the classifier, a five-channel analysis was performed. The Cz channel and four surrounding channels (Pz, C3, C4 and Fz) were considered for the ablation study. Table \ref{AblationResults} shows results for the classification performance with a reduced number of channels.

\subsection*{Classification Performance - Case 3: Delta-Modulated Spike Train Input Data}

The effect of processing the input EEG data from all 19 channels into spike train data before passing to the CSNN network was also studied, namely its impact on classification performance of the network. The filtered and segmented EEG data, normalized to lie in the range of 0 to 1 inclusively, was transformed into a 19 channel array of spike train data by monitoring the change in value of successive data points in each channel. If the value change was greater than a threshold value, a spike '1' was recorded. If not, a '0' was recorded. The result was a binary array of the same dimensions as the floating-point input. The threshold value was varied from 0.05 to 1 which resulted in spike trains of differing densities. The CSNN model used had the same parameters as in Case 1. 

Table \ref{spikeTrainRes} shows the 10-fold classification results for Case 3. The results indicate that good predictive performance can still be achieved when the floating-point EEG data were converted into spike trains prior to being input to the network, if a suitable threshold value was selected. Spike train conversion thresholds that are too small or too large constrain the abilities of the CSNN to learn effectively, most likely because features important to the correct classification of data were being obscured. If the threshold was too small, then the spike train becomes saturated, making it difficult for the network to determine the important features. If the threshold was too large, then important features may not be captured at all.  

The best results were obtained with a threshold of 0.5. The sensitivity of the classification results to the threshold value was studied using two-sample t-tests comparing the classification measures corresponding to the 0.5 threshold with the measures corresponding to the other thresholds. The t-test results shown in Table \ref{spikeTrainRes} indicate significant performance degradation above a threshold of 0.625 and below 0.375. This implies that a range of threshold values could be used to obtain statistically similar results, offering flexibility in the threshold selection. The performance of the CSNN using a threshold value of 0.5 was comparable to the EEGNet results when trained on the floating-point data. A two-sample t-test showed that TNR for the CSNN was statistically better (\emph{p}-value was 0.041), but the accuracy and F1-score metrics were statistically similar (\emph{p}-values were 0.700 and 0.537, respectively). On the other hand, the TPR for EEGNet was significantly better (\emph{p}-value was 0.028).

\begin{table*}[!t]
\centering
\caption{Classification performance with delta modulated spike train input data (best performance in each classification measure highlighted in bold font)}
\resizebox{\textwidth}{!} 
{ 
\begin{tabular}{c c c c c c c c c c c}
\hline 
\\[-1.75ex]
\multirow{2}{*}{Threshold} & \multicolumn{2}{c}{Accuracy (\%)} & \multicolumn{2}{c}{TPR (\%)} & \multicolumn{2}{c}{TNR (\%)} & \multicolumn{2}{c}{F1-score} & Epochs
\\[0.5ex]
& Mean (SD) & \emph{p}-value & Mean (SD) & \emph{p}-value & Mean (SD) & \emph{p}-value & Mean (SD) & \emph{p}-value & Mean (SD)
\\[0.75ex]
\hline\hline
\\[-1.5ex]
0.05 & 90.43 (0.339) & $<$ 0.001 & 69.74 (2.60) & $<$ 0.001 & 95.57 (0.53) & $<$ 0.001 & 0.74 (0.01) & $<$ 0.001 & 286 (100)
\\[1ex]

0.25 & 94.57 (0.65) & $<$ 0.001 & 82.61 (2.29) & $<$ 0.001 & 97.54 (0.71) & $<$ 0.001 & 0.86 (0.02) & $<$ 0.001 & 208 (38)
\\[1ex]
0.375 & 96.89 (0.49) & 0.021 & 89.38 (2.54) & 0.035 & 98.75 (0.37) & 0.098 & 0.92 (0.01) & 0.020 & 243 (51)
\\[1ex]

0.5 & $\textbf{97.56}$ (0.43) & $N/A^{\dagger}$ & $\textbf{91.64}$ (1.98) & $N/A^{\dagger}$ & $\textbf{99.03}$ (0.23) & $N/A^{\dagger}$ & \textbf{0.94} (0.01) & $N/A^{\dagger}$ & 270 (45) \\[1ex]

0.625 & 97.27 (0.55) & 0.150 & 91.26 (2.06) & 0.646 & 98.76 (0.47) & 0.142 & 0.93 (0.02) & 0.165 & 184 (40) \\[1ex]

0.75 & 95.80 (0.62) & $<$ 0.001 & 85.95 (3.19) & $<$ 0.001 & 98.25 (0.44) & $<$ 0.001 & 0.89 (0.02) & $<$ 0.001 & 261 (77)\\[1ex]

1 & 88.83 (1.13) & $<$ 0.001 & 63.77 (4.79) & $<$ 0.001 & 95.05 (0.83) & $<$ 0.001 & 0.69 (0.04) & $<$ 0.001 & 208 (49)
\\[1ex]
\hline
\\[-2ex]
$^\dagger$ \text{Threshold Tested}
\end{tabular}
}
\label{spikeTrainRes}
\end{table*}

\section*{Discussion \label{sec: discussion}}
The results presented here show that the CSNN can be used as a classifier for detecting features in EEG data that predict braking intention, which occurs before the actual physical activity. To benchmark the CSNN performance, results were compared to a standard CNN, EEGNet and three GNN models using a 10-fold cross-validation scheme with the CSNN achieving the highest performance and with more consistency. The \emph{p}-values from two-sample t-tests in Table \ref{StandardResults} show a significantly higher performance of the CSNN over the GNNs in almost every metric category (except TNR of the GCN network, where the \emph{p}-value was slightly above 0.05). This result is not surprising when the means of the metrics are compared. This fact is in stark contrast to the \emph{p}-values of the CNN and EEGNet, where the CNN has noticeably lower mean values than the CSNN and is statistically similar; however, EEGNet is the closest of any of the other models and is statistically different. This can be explained by the fact that the CNN had enough folds containing results that were nearly identical to the CSNN performance, but also had a couple of folds with poor performance that biased the grand averages. As a result, the overall model performance was not significantly different than the CSNN. On the other hand, although EEGNet performance was competitive and consistent, it did not quite match the CSNN performance on any fold. Therefore, the \emph{p}-values indicated statistically significant higher performance of the CSNN compared with EEGNet. The authors hypothesize that a possible explanation for the CSNN's success is that converting the floating-point numbers to spike trains allows it to filter more efficiently, passing the most important feature maps to the next layer. 

Despite the CSNN's success in classification, it had the longest average training time and largest inference time by a large margin. While computational efficiency could be improved by deploying CSNNs on neuromorphic hardware, those gains were not realized in this study where only a von Neumann computer was utilized. Neuromorphic chips, such as the Intel Loihi chip, have been shown to produce faster training times than von Neumann processors and deep learning accelerators\cite{davies_wild_orchard_sandamirskaya_guerra_joshi_plank_risbud_2021}. 

Results of the five-channel ablation study indicate that a few strategically chosen channels may be sufficient with the CSNN's classification performance being nearly identical to that attained with all 19 channels. It can also be seen that the CNN performance increased considerably compared to the results shown in Table \ref{StandardResults} with all 19 channels. Although the mean performance of the CNN was substantially lower when all 19 channels were used, the \emph{p}-values in Table \ref{StandardResults} indicate that the results were not significantly different from the CSNN. As noted previously, a couple of folds with poor performance biased the CNN grand average. Therefore, the boost in CNN performance to the level of CSNN in the ablation study is not altogether surprising and could simply have resulted from more consistent performance across folds when using only five channels.

The spike train conversion results shown in Table \ref{spikeTrainRes} have significant implications because converting the floating-point EEG data to spike trains at the outset could allow for additional energy savings, thereby taking full advantage of any neuromorphic computing hardware used to implement the CSNN for large, complex datasets and/or in real-time applications. Findings from this study can be exploited in future work to implement the CSNN on a neuromorphic platform to study the actual energy efficiency and feasibility for on-line learning in real-time applications. 

The Cz grand average calculation and the analysis itself is based on the assumption of zero phase shift of the CNV pattern relative to the external stimuli.  Lew \emph{et al} \cite{10.3389/fneng.2012.00013} and Khaliliardali \emph{et al} \cite{khaliliardali_chavarriaga_gheorghe_millan_2015} discuss in great detail SCPs, of which CNVs are a subset. In these studies, the SCP was determined to begin as early as 1.5 seconds before the onset of movement and therefore the crucial aspects of the CNV should be contained within the data between the "1" count marker and the stop marker. Variations of the exact timing of the pattern will occur between trials but as mentioned above, the grand average pattern obtained in this study are consistent with the results reported in the literature.
Also, it is well known that EEG has poor spatial resolution. For this reason, the CNV pattern in Figure \ref{eeg_signal} appears to occur across the entire brain instead of the central area as expected. Results with a higher number of channels than the 19 channels considered in this study have reported greater regional localization \cite{khaliliardali_chavarriaga_gheorghe_millan_2015}. Furthermore, ablation study results presented in Table \ref{AblationResults} confirm that the CNV is occurring in the central area.

The EEG data was collected from 15 participants with the dataset containing a total of 3244 trials that were cleaned and segmented into 10,702 data segments. Although the number of participants in this study is relatively small, it is on par with other studies in the literature. The classification experiment conducted in-house involved using a simulated-realistic testbed with the participant operating a remote-controlled vehicle using a live video feed under ideal conditions. It would be interesting to study the performance of the CSNN when the participants are cognitively stressed, 
for example, when under fatigue or in the presence of distractions. Also of interest would be to study the abilities of the CSNN in other EEG-BCI applications such as P300, motor imagery, motor-related cortical potentials and steady-state evoked potentials or to explore the use of CSNNs for movement intention detection using Bereitschaftpotentials, which are well-known as an indicator of movement preparation within the brain, much like CNVs. Their use in braking intention, or other driving related task intention, using CSNNs would be an interesting research topic to explore.

\section*{Methods \label{sec: NN models}}

\begin{figure*}[!t]
\centering
\includegraphics[width=\textwidth]{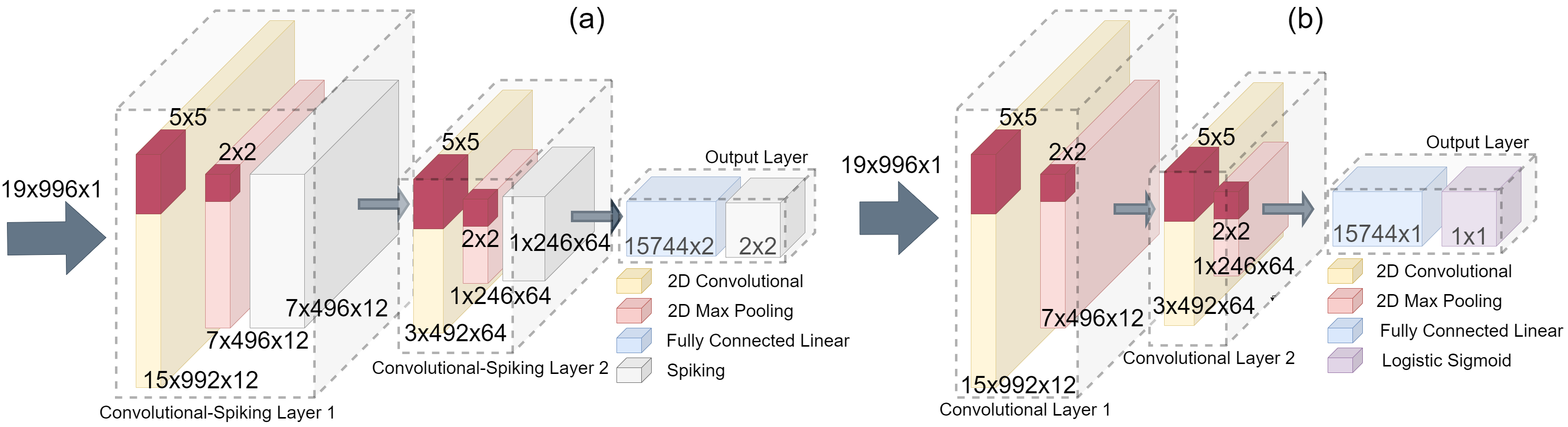}
\caption{Schematic of neural network architectures. (a) CSNN. (b) CNN.}
\label{csnnDia}
\end{figure*}

\begin{figure*}[!t]
\centering
\includegraphics[width = \textwidth]{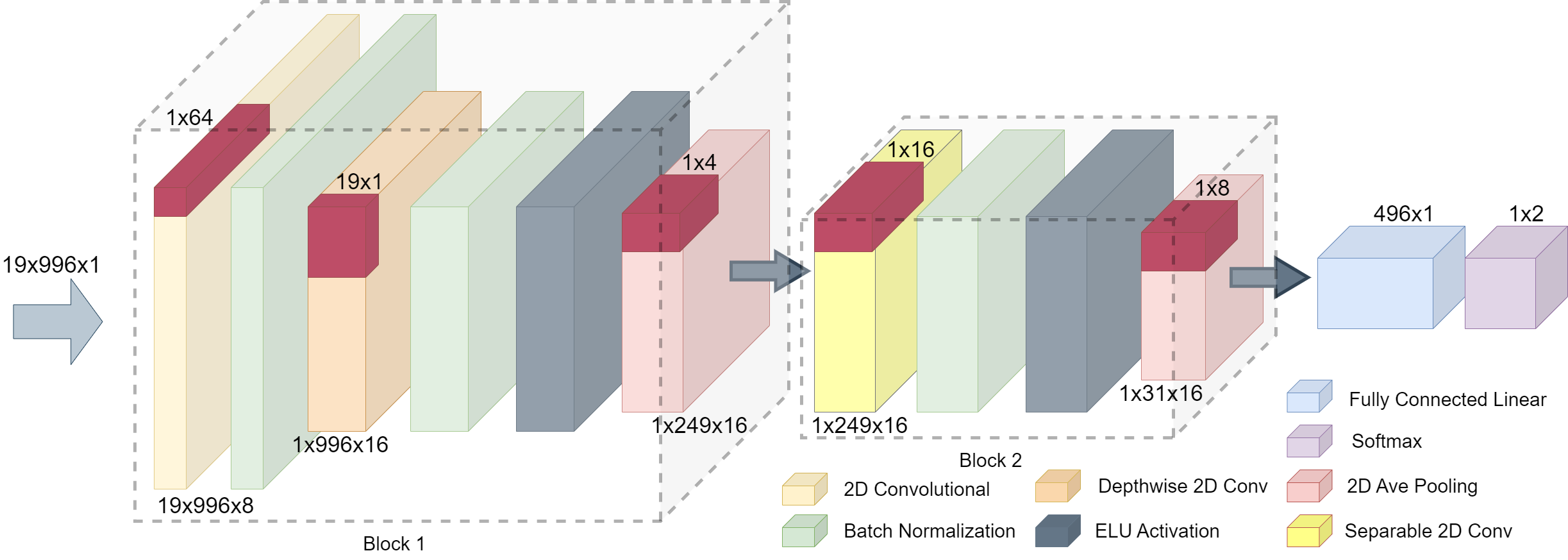}
\caption{Schematic diagram of EEGNet.}
\label{eegnet}
\end{figure*}

\subsection*{Convolutional Spiking Neural Networks}

CSNNs are deep networks comprised of standard convolutional layers that extract feature maps from the input data before passing these feature maps to subsequent spiking layers. These combination layers are referred to here as "convolutional-spiking" layers. For classification, the output layer is composed of a fully connected layer with linear activation function followed by a spiking layer. Figure \ref{csnnDia}(a) shows a schematic of the specific CSNN used in this study. It is comprised of two convolutional-spiking layers followed by the output layer. Each convolutional-spiking layer is comprised of a two-dimensional convolution layer followed by a two-dimensional max pooling layer and ending with an output spiking layer. The spiking layer in each convolutional-spiking layer is composed of a tensor of LIF neurons having the same shape as the shape of the input to the layer. In the output layer, the fully connected layer and the subsequent output LIF spiking layer both have two neurons for the two classes in the EEG dataset. The predicted output of the CSNN was determined by counting the number of spikes output by these two neurons and setting the predicted label to the class represented by the neuron which produced the most spikes. A tie would result in a predicted class of '0'.

Due to the non-differentiable nature of the output of spiking neurons, training SNNs is difficult and requires special approaches beyond simply using standard backpropagation. If the spiking behavior of a neuron is represented as:

\begin{equation}
\label{spTwo}
S[t] = \Theta(U_{mem}[t]-\theta)
\end{equation}

\noindent
where $\Theta(\cdot)$ is the heavy-side function, $U_{mem}[t]$ is the membrane potential of the neuron and $\theta$ is the spiking threshold, the derivative of (\ref{spTwo}) with respect to $U_{mem}$ is the dirac delta function:

\begin{equation}\label{dirac}
    \frac{\partial S}{\partial U_{mem}[t]}=\delta (U_{mem}[t] - \theta) \in \{0,\infty\}
\end{equation}

\noindent
which is defined as zero for all time except where $U_{mem} = \theta$ at which it is infinity. This leads to the "dead neuron" problem for training using backpropagation. To mitigate this, the surrogate gradient approach is employed wherein during the "backward-pass", when the gradient of the loss function due to the network parameters is being computed, the heavy-side function is approximated using a sigmoidal function, thereby creating a readily differential function. The exact function used as the surrogate in this paper is described as:

\begin{equation}\label{surro}
    \tilde{S} = \frac{U_{mem}[t]-\theta}{1+k|U_{mem}[t]-\theta|}
\end{equation}

\noindent 
where $k$ is known as the 'slope' and determines the smoothness of the surrogate function. The derivative of (\ref{surro}) is then obtained as:

\begin{equation}\label{surroDer}
\frac{\partial \tilde{S}}{\partial U_{mem}[t]} = \frac{1}{(k|U_{mem}[t]-\theta|+1)^2}   
\end{equation}

\noindent
From (\ref{surroDer}) it can be seen that as k increases, (\ref{surroDer}) converges to (\ref{dirac}). 
For a more detailed explanation of SNNs and their training, the reader is referred to \cite{eshraghian2021training}.

\subsection*{Convolutional Neural Network}

For the sake of a direct spiking vs. non-spiking comparison, a CNN composed of a largely identical architecture as that of the CSNN is considered to fully quantify any differences in performance that may arise by the addition of the spiking layers. As shown in Figure \ref{csnnDia}(b), the CNN has two convolutional layers, each including a max pooling layer, followed by a fully connected linear layer and ending with a logistic sigmoid output layer for class prediction.

\subsection*{EEGNet}
EEGNet is a single CNN architecture designed for classification tasks across multiple EEG-based BCI domains (P300 visual-evoked potentials, error-related negativity responses, movement-related cortical potentials and sensory motor rhythms)\cite{lawhern_2018}. It consists of two blocks of convolutional layers followed by a dense layer with finally a softmax layer. It is compact in terms of the number of model parameters (see Figure \ref{eegnet}). In the first block, two convolutional operations are done in sequence. This block starts with a temporal convolution to learn frequency filters followed by "depthwise" convolution to learn frequency-specific spatial filters. The second block also includes two convolutional operations. The first is another "depthwise" convolution to individually learn the temporal feature map and the second is "pointwise" convolution to optimally combine the feature maps. These two convolutions are combined into one layer, termed "Separable 2D Convolution". The output of the second block of layers is then flattened and passed to the dense and softmax layer for generation of the predicted class.

\subsection*{Graph Neural Networks}

\begin{figure*}[!t]
\centering
\includegraphics[width = \textwidth]{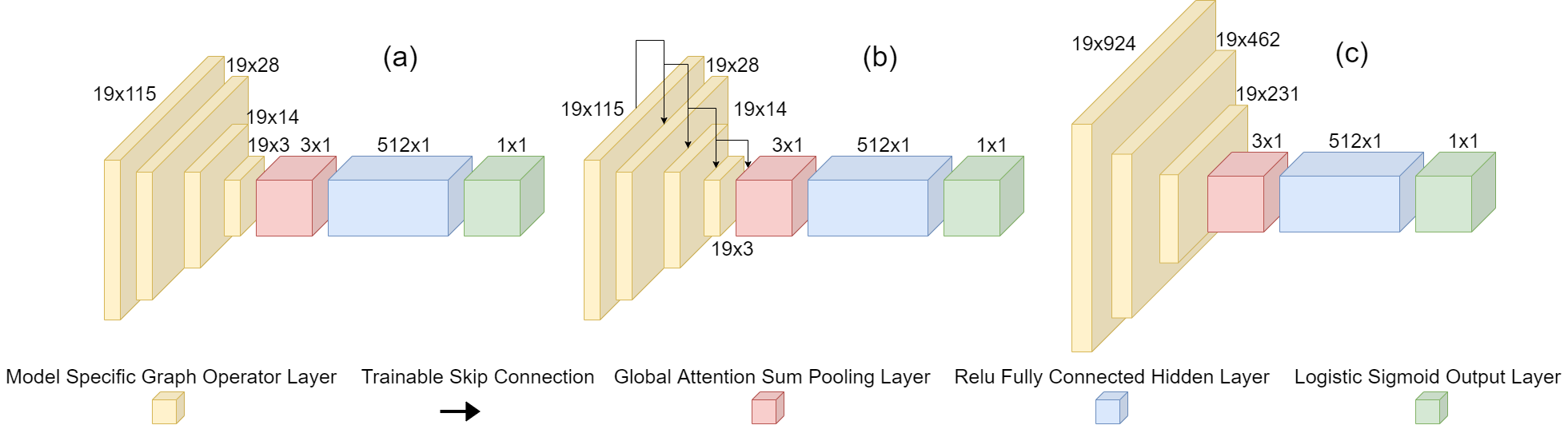}
\caption{Schematic diagrams of GNNs. (a) GCNConv. (b) GCSConv. (c) GINConv.}
\label{gcngcs}
\end{figure*}

GNNs are a specialized version of neural networks designed to operate on graph data. A graph is a grouping of data with defined internal relationships (edges) between objects (nodes) where these relationships may or may not be euclidean in nature. Mathematically, a graph is typically represented as: $\mathcal{G}=(\mathcal{V},\mathcal{E},A)$ where $\mathcal{V}$ represents a finite set of nodes having length $|\mathcal{V}| = N$, $\mathcal{E}$ is a set of edges between the nodes, and $A \in \mathcal{R}^{N \times N}$ is the adjacency matrix containing the edge weights. Graph data is input to a GNN as a matrix of node feature vectors: $X \in \mathbf{R}^{N \times n}$, where $N$ is the number of nodes and $n$ is the number of node features, along with its adjacency matrix, $A$, and sometimes a set of edge features, $E$. Some graph operations also include the diagonal degree matrix $D$ where $D_{ii} = \sum_{j} A_{ij}$. For more details on graph theory and a detailed survey providing a comprehensive overview of GNNs, the reader is referred to \cite{wilson_2015, wu_pan_chen_long_zhang_yu_2021}.
Because of the spatial relationship between electrodes, EEG data can naturally be represented as graphs with each data input sharing the same node and edge structure, thereby differing only in node data. Expressing the data as graphs allows for the non-euclidean spatial relationships to be exploited as extra information available to the classifier. For this reason, EEG-BCI applications have used GNNs previously, to notable effect.

As shown in Figure \ref{gcngcs}, the three GNN architectures used in this work differ only in their initial graph processing layers, whilst universally sharing the last three layers. Each network possessed a global attention sum pool graph aggregation layer which feeds into a fully-connected hidden layer with rectified linear unit (ReLu) activation. The final layer is a classification output layer consisting of one neuron with a logistic sigmoid activation function. The global attention sum pool layer computes

\begin{equation}
X^{'}= \sum_{i=1}^{N}\alpha_{i}X_{i}, \quad \alpha = \text{softmax}(Xa)
\end{equation}

\noindent
where $a$ is a trainable weight vector, $X$ is the layer input tensor, $N$ is the number of nodes in the input graph, and the softmax operation is applied over nodes instead of features. 

As shown in Figure \ref{gcngcs}a, the first architecture, GCNConv network \cite{https://doi.org/10.48550/arxiv.1609.02907}, consists of four graph convolutional (GCN) layers. 
The GCN layers perform the operation:

\begin{equation}
    Y = \hat{D}^{-1/2}\hat{A}\hat{D}^{-1/2}XW+b
\end{equation}

\noindent
where $Y$ is the output of the layer, $\hat{A} = A + I$ is the adjacency matrix of the input graph plus the identity matrix of appropriate shape, $\hat{D} = \sum_{j} \hat{A}_{ij}$ is the degree matrix,  $X$ is the layer input tensor, $W$ is the layer weights, and $b$ is the layer matrix.

Figure \ref{gcngcs}b shows the second architecture, which is the GCSConv network. This consists of 4 GCS layers, which are GCN layers with an added, trainable skip connection.
The GCS layer operation is described by:

\begin{equation}
    Y = D^{-1/2}AD^{-1/2}XW_{1} + XW_{2} + b
\end{equation}

\noindent
where $Y$ is the output of the layer, $D$ is the degree matrix, $A$ is the adjacency matrix, $X$ is the node feature matrix, $W_{1}$ and $W_{2}$ are the two sets of layer weights, and $b$ is the layer bias.

The third architecture is the GIN architecture \cite{https://doi.org/10.48550/arxiv.1810.00826} and is shown in Figure \ref{gcngcs}c. This architecture contains three graph isometric network (GIN) layers where each layer performs the following operation for each node in the input matrix:

\begin{equation}
Y_{i} = MLP\left( (1+\epsilon) \cdot x_{i} + \sum_{j \in N(i)} x_{j}\right)
\end{equation}

\noindent
where $MLP(\cdot)$ is a multi-layer perceptron, $\epsilon$ is a learned parameter and $x_{i}$ is the $i^{th}$ node of the input matrix. 

The adjacency matrix, which would be common to all data samples, for all the three GNNs, was calculated as follows:

\begin{equation}
A = |P| - I
\end{equation}

\noindent
where $A \in \mathbf{R}^{N \times N}$ is the adjacency matrix, $|P| \in \mathbf{R}^{N \times N}_{+}$ is the absolute value of the Pearson's correlation coefficient of the dataset and $I^{N \times N}$ is the identity matrix.

\subsection*{Experimental Design}\label{exp_design}

The total number of participants in the experiment was 15 (13 male, 2 female). They consisted of Missouri University of Science and Technology students and professors, all in healthy condition. The participants had normal or corrected to normal vision and had normal hearing. The experiment received approval from the University of Missouri Institutional Review Board, and all experiments were performed in accordance with relevant guidelines and regulations. Written informed consent was obtained from all subjects and/or their legal guardian(s) prior to their participation. Further, written informed consent was obtained for publication of identifying information/images in an online open-access publication.

\begin{figure*}[!t]
\centering
\includegraphics[width=\textwidth]{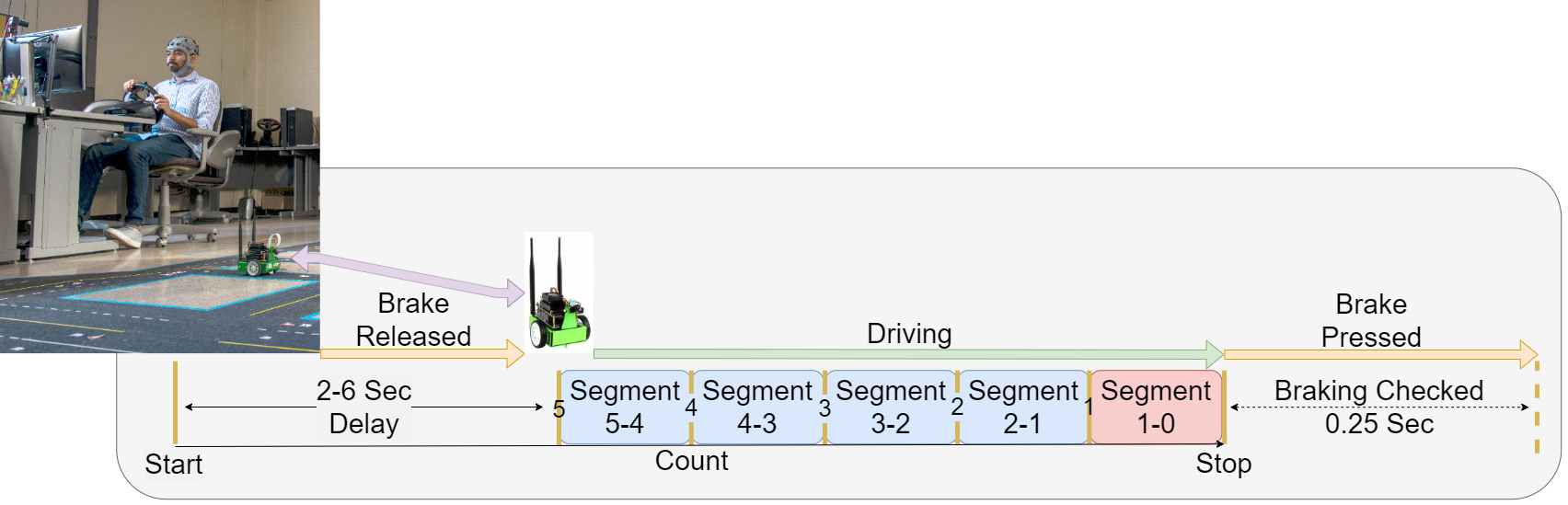}
\caption{Experimental design illustration of a trial. Photo by Micheal Pierce/Missouri S\&T}
\label{ExpDes}
\end{figure*}

The objective of the experiment was to induce a predictable response in the participants such that any anticipatory signals that may occur can be reliably measured and recorded using an EEG. The experiment simulated a real-world driving environment wherein the participants operated an open-source remote-controlled robot called JetBot (built using Waveshare's Jetbot AI Kit and Nvidia's 4GB Jetson-Nano) on a novel testbed designed to simulate urban roadways (see picture inserted in Figure \ref{ExpDes}) .
The testbed boundary was marked with standard masking tape and the track material was Delxo's anti-slip tape (with 80-grit granularity) to provide additional traction to the JetBot wheels. The participants navigated the JetBot in the testbed lanes using a Logitech G29 Driving Force racing wheel and pedal setup, while watching a live video feed cast to a computer monitor from an on-board camera. 
The JetBot was programmed to drive at a constant speed without the participant pressing the acceleration pedal, necessitating only the use of the steering wheel and brake pedal for full control. There were no other 'vehicles' or obstructions on the testbed lanes and participants were free to navigate anywhere within the testbed boundaries. 

The EEG signals of the participants were recorded using a Neuroelectrics ENOBIO 20 EEG headset. The electrode setup used was the international 10-20 standard and the sampling frequency during the experiment was 500 Hz. Data was collected from 19 channels by applying a high conductivity Signagel saline gel on the electrodes to increase the quality of data capture. The data acquisition software used was the Neuroelectrics NIC2 software, which featured its own EEG signal quality monitor. The quality monitor assessed the EEG signal by computing a quality index (QI) that was dependent on: i) line noise, which was defined as electrical noise originating from surrounding power lines; ii) main noise, which was defined as the signal power of the standard EEG band; and iii) offset, which was the mean value of the waveform. Specifically, $QI$ was calculated as \cite{Neuroelectrics}:

\begin{equation}
QI(t) = \tanh\Bigg(\sqrt{\bigg(\frac{\zeta_{L}(t)}{W_{\zeta_{L}}}\bigg)^2 + \bigg(\frac{\zeta_{m}(t)}{W_{\zeta_{m}}}\bigg)^2 + \bigg(\frac{O(t)}{W_{O}}\bigg)^2}\Bigg)
\end{equation}

\noindent
where $\zeta_{L}(t)$ and $W_{\zeta_{L}}$ denote the line noise and line noise normalizing weight (= 100 $\mu$V), respectively, $\zeta_{m}(t)$ and $W_{\zeta_{m}}$ denote the main noise and main noise normalizing weight (= 250 $\mu$V), respectively, and $O(t)$ and $W_O$ denote the offset and offset normalizing weight (= 280 mV), respectively.
The NIC2 software indicators used a color scheme to indicate different levels of QI. A green indicator meant that the signal had a QI between 0 and 0.5, an orange indicator meant a QI of between 0.5 and 0.8, and a red indicator meant a QI of 0.8 to 1. For this experiment, green indicators for all channels was the standard; however, brief periods of orange indicator were considered acceptable. The data was filtered using a 60 Hz filter during capture to help reduce electrical noise and all channels were captured with reference to the Common Mode Sense channel (20th channel) which was fixed to the participant's right ear lobe.

The experimental design was based, in part, on the experiment conducted in \cite{khaliliardali_chavarriaga_gheorghe_millan_2015} and is illustrated in Figure \ref{ExpDes}. Each participant underwent 8 sets of 30 trials each for a total of 240 trials, with short 5-10 minute breaks between sets. Each trial consisted of a set of audible commands issued by MATLAB that included a "Start" command, upon hearing which the participant would release the brake allowing the JetBot to move, followed by a countdown from 5 to 1 and ending with a "Stop" command, when the participant would immediately stop the JetBot by pressing the brake. To ensure that the participants responded in a timely fashion, activity at the brake pedal was monitored and any trial where the brake pedal depression did not register a numerical reading higher than 0.05\% of its total depression range within 0.25 seconds of the issuance of the "Stop" command were marked for removal. Trials where the participant braked too early were manually marked for removal as well. The EEG recording would begin concurrently with the issuance of the "Start" command, markers corresponding to the countdown numbers would be applied to the data concurrently with each audio count and the EEG recording would stop at detection of braking action or after the 0.25 second delay to check brake pedal depression.

\subsection*{Data Preprocessing}\label{data_proc}

\begin{figure*}[!t]
\centering
\includegraphics[width = \textwidth]{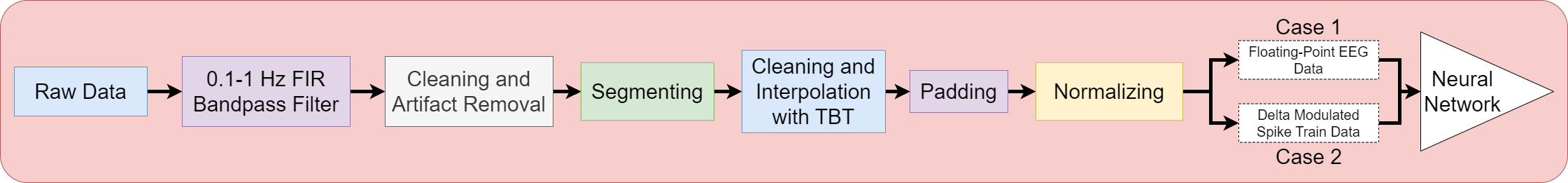}
\caption{Data processing pipeline.}
\label{DP}
\end{figure*}

The data was processed in several steps using the open source EEG toolkit EEGLAB \cite{delorme_makeig_2004} along with the TBT plugin \cite{mattan_s_ben_shachar_2022_5948294} for only the trials that were not marked for removal during the experiments (see Figure \ref{DP}). 

\begin{enumerate}[leftmargin=3ex]
\item Each trial was:
\begin{enumerate}[leftmargin=3.5ex]
    \item Spectrally filtered using a FIR bandpass filter from 0.1 Hz to 1 Hz as suggested in \cite{garipelli_chavarriaga_del_r_millan_2013}. 
    \item Cleaned using EEGLAB's built-in automated cleaning function "Clean\_RawData and ASR" (Artificial Subspace Reconstruction) to remove bad channels under the following criteria: If the channel: i) was flat for more than five seconds; ii) correlated at less than 0.8 to an estimate based on nearby channels; and iii) contained more than four standard deviations of line noise relative to its signal. The ASR attempted to correct bad data periods containing artifacts and its maximum acceptable 0.5 second window standard deviation limit was set to a conservative 20 standard deviations. The values used in this step were the standard default values in EEGLAB and were also used as part of a pre-processing scheme in a previous study \cite{bagdasarov_roberts_brechet_brunet_michel_gaffrey_2022}. \item Segmented by slicing the data according to the markers corresponding to the countdown numbers or the "Stop" command. For example, as shown in Figure \ref{ExpDes}, the first segment consisted of taking only the data that occurred between the "5" count marker and the "4" count marker. The segments from 5 to 1 were given a "0" label or were regarded as not containing an intention signal and the segment between "1" and "Stop" was labelled as "1" and was regarded as containing the signal of interest. Each segment was then baseline corrected by subtracting the mean value of the segment from every value in the segment.
    \end{enumerate}

\item Each segment was then further cleaned using the TBT plugin to remove high amplitude noise. Channels were removed from the segment if either of the following two criteria was met for a data period duration of 10\% of an segment or more \cite{bagdasarov_roberts_brechet_brunet_michel_gaffrey_2022}: i) if they exceeded $\pm 100$ $\mu$v in magnitude, or ii) if the joint probabilities (i.e., probabilities of activity) exceeded 3 standard deviations for local or global thresholds. If either criterion was met for less than 10\% of an segment, then the offending data period was removed and subsequently interpolated. Segments with more than 50\% of channels removed (i.e., 10 channels) were omitted entirely. Any removed channels were re-interpolated after the cleaning was finished for consistency in input dimension. 

\item Every data point was padded to a uniform length of 1848, the size of the largest dataset in a trial, by appending zeros to the end. Padding data should not significantly change the prediction results and is commonly done when using machine learning algorithms on datasets containing images having different sizes. Furthermore, padding the data should less significantly alter the dataset versus truncating data, which allows for the possibility of deletion of key features.  

\item Finally, each channel was normalized such that the data lay within a range of 0 to 1. The particular equation used was:

\begin{equation}
\overline{X}_{i} = \frac{X_{i} - \min(X_{i})}{\max(X_{i}) - \min(X_{i})}
\end{equation}

\noindent
where $i$ is the $i^{th}$ data channel and $X$ is the data vector corresponding to that channel, $\max(\cdot)$ and $\min(\cdot)$ are the maximum and minimum channel values per segment, respectively.
    
\end{enumerate}

\subsection*{Data Availability}\label{data_avail}

\noindent The datasets used and/or analyzed during the current study are available from the corresponding author on reasonable request.


\begin{thebibliography}{10}
\urlstyle{rm}
\expandafter\ifx\csname url\endcsname\relax
  \def\url#1{\texttt{#1}}\fi
\expandafter\ifx\csname urlprefix\endcsname\relax\def\urlprefix{URL }\fi
\expandafter\ifx\csname doiprefix\endcsname\relax\def\doiprefix{DOI: }\fi
\providecommand{\bibinfo}[2]{#2}
\providecommand{\eprint}[2][]{\url{#2}}

\bibitem{Christensen_2022}
\bibinfo{author}{Christensen, D.~V.} \emph{et~al.}
\newblock \bibinfo{journal}{\bibinfo{title}{2022 roadmap on neuromorphic
  computing and engineering}}.
\newblock {\emph{\JournalTitle{Neuromorphic Computing and Engineering}}}
  \textbf{\bibinfo{volume}{2}}, \bibinfo{pages}{022501},
  \doiprefix\url{10.1088/2634-4386/ac4a83} (\bibinfo{year}{2022}).

\bibitem{tan_sarlija_kasabov_2020}
\bibinfo{author}{Tan, C.}, \bibinfo{author}{Sarlija, M.} \&
  \bibinfo{author}{Kasabov, N.}
\newblock \bibinfo{journal}{\bibinfo{title}{Spiking neural networks:
  Background, recent development and the neucube architecture}}.
\newblock {\emph{\JournalTitle{Neural Processing Letters}}}
  \textbf{\bibinfo{volume}{52}}, \bibinfo{pages}{1675--1701},
  \doiprefix\url{10.1007/s11063-020-10322-8} (\bibinfo{year}{2020}).

\bibitem{dora_kasabov_2021}
\bibinfo{author}{Dora, S.} \& \bibinfo{author}{Kasabov, N.}
\newblock \bibinfo{journal}{\bibinfo{title}{Spiking neural networks for
  computational intelligence: An overview}}.
\newblock {\emph{\JournalTitle{Big Data and Cognitive Computing}}}
  \textbf{\bibinfo{volume}{5}}, \bibinfo{pages}{67},
  \doiprefix\url{10.3390/bdcc5040067} (\bibinfo{year}{2021}).

\bibitem{tavanaei_ghodrati_kheradpisheh_masquelier_maida_2019}
\bibinfo{author}{Tavanaei, A.}, \bibinfo{author}{Ghodrati, M.},
  \bibinfo{author}{Kheradpisheh, S.~R.}, \bibinfo{author}{Masquelier, T.} \&
  \bibinfo{author}{Maida, A.}
\newblock \bibinfo{journal}{\bibinfo{title}{Deep learning in spiking neural
  networks}}.
\newblock {\emph{\JournalTitle{Neural Networks}}}
  \textbf{\bibinfo{volume}{111}}, \bibinfo{pages}{47--63},
  \doiprefix\url{10.1016/j.neunet.2018.12.002} (\bibinfo{year}{2019}).

\bibitem{davies_wild_orchard_sandamirskaya_guerra_joshi_plank_risbud_2021}
\bibinfo{author}{Davies, M.} \emph{et~al.}
\newblock \bibinfo{journal}{\bibinfo{title}{Advancing neuromorphic computing
  with loihi: A survey of results and outlook}}.
\newblock {\emph{\JournalTitle{Proceedings of the IEEE}}}
  \textbf{\bibinfo{volume}{109}}, \bibinfo{pages}{911--934},
  \doiprefix\url{10.1109/jproc.2021.3067593} (\bibinfo{year}{2021}).

\bibitem{8354825}
\bibinfo{author}{Lee, C.}, \bibinfo{author}{Srinivasan, G.},
  \bibinfo{author}{Panda, P.} \& \bibinfo{author}{Roy, K.}
\newblock \bibinfo{journal}{\bibinfo{title}{Deep spiking convolutional neural
  network trained with unsupervised spike-timing-dependent plasticity}}.
\newblock {\emph{\JournalTitle{IEEE Transactions on Cognitive and Developmental
  Systems}}} \textbf{\bibinfo{volume}{11}}, \bibinfo{pages}{384--394},
  \doiprefix\url{10.1109/TCDS.2018.2833071} (\bibinfo{year}{2019}).

\bibitem{vaila_chiasson_saxena_2019}
\bibinfo{author}{Vaila, R.}, \bibinfo{author}{Chiasson, J.} \&
  \bibinfo{author}{Saxena, V.}
\newblock \bibinfo{journal}{\bibinfo{title}{Feature extraction using spiking
  convolutional neural networks}}.
\newblock {\emph{\JournalTitle{Proceedings of the International Conference on
  Neuromorphic Systems}}} \doiprefix\url{10.1145/3354265.3354279}
  (\bibinfo{year}{2019}).

\bibitem{lee_panda_srinivasan_roy_2018}
\bibinfo{author}{Lee, C.}, \bibinfo{author}{Panda, P.},
  \bibinfo{author}{Srinivasan, G.} \& \bibinfo{author}{Roy, K.}
\newblock \bibinfo{journal}{\bibinfo{title}{Training deep spiking convolutional
  neural networks with stdp-based unsupervised pre-training followed by
  supervised fine-tuning}}.
\newblock {\emph{\JournalTitle{Frontiers in Neuroscience}}}
  \textbf{\bibinfo{volume}{12}}, \doiprefix\url{10.3389/fnins.2018.00435}
  (\bibinfo{year}{2018}).

\bibitem{barchid_mennesson_djeraba_2021}
\bibinfo{author}{Barchid, S.}, \bibinfo{author}{Mennesson, J.} \&
  \bibinfo{author}{Djeraba, C.}
\newblock \bibinfo{journal}{\bibinfo{title}{Deep spiking convolutional neural
  network for single object localization based on deep continuous local
  learning}}.
\newblock {\emph{\JournalTitle{2021 International Conference on Content-Based
  Multimedia Indexing (CBMI)}}} \doiprefix\url{10.1109/cbmi50038.2021.9461880}
  (\bibinfo{year}{2021}).

\bibitem{1198140}
\bibinfo{author}{Matsugu, M.}, \bibinfo{author}{Mori, K.},
  \bibinfo{author}{Ishii, M.} \& \bibinfo{author}{Mitarai, Y.}
\newblock \bibinfo{title}{Convolutional spiking neural network model for robust
  face detection}.
\newblock In \emph{\bibinfo{booktitle}{Proceedings of the 9th International
  Conference on Neural Information Processing, 2002. ICONIP '02.}},
  vol.~\bibinfo{volume}{2}, \bibinfo{pages}{660--664 vol.2},
  \doiprefix\url{10.1109/ICONIP.2002.1198140} (\bibinfo{year}{2002}).

\bibitem{dong_huang_xu_2018}
\bibinfo{author}{Dong, M.}, \bibinfo{author}{Huang, X.} \& \bibinfo{author}{Xu,
  B.}
\newblock \bibinfo{journal}{\bibinfo{title}{Unsupervised speech recognition
  through spike-timing-dependent plasticity in a convolutional spiking neural
  network}}.
\newblock {\emph{\JournalTitle{PLOS ONE}}} \textbf{\bibinfo{volume}{13}},
  \doiprefix\url{10.1371/journal.pone.0204596} (\bibinfo{year}{2018}).

\bibitem{xing_di_caterina_soraghan_2020}
\bibinfo{author}{Xing, Y.}, \bibinfo{author}{Di~Caterina, G.} \&
  \bibinfo{author}{Soraghan, J.}
\newblock \bibinfo{journal}{\bibinfo{title}{A new spiking convolutional
  recurrent neural network (scrnn) with applications to event-based hand
  gesture recognition}}.
\newblock {\emph{\JournalTitle{Frontiers in Neuroscience}}}
  \textbf{\bibinfo{volume}{14}}, \doiprefix\url{10.3389/fnins.2020.590164}
  (\bibinfo{year}{2020}).

\bibitem{turkson_qu_mawuli_eghan_2021}
\bibinfo{author}{Turkson, R.~E.}, \bibinfo{author}{Qu, H.},
  \bibinfo{author}{Mawuli, C.~B.} \& \bibinfo{author}{Eghan, M.~J.}
\newblock \bibinfo{journal}{\bibinfo{title}{Classification of alzheimer’s
  disease using deep convolutional spiking neural network}}.
\newblock {\emph{\JournalTitle{Neural Processing Letters}}}
  \textbf{\bibinfo{volume}{53}}, \bibinfo{pages}{2649--2663},
  \doiprefix\url{10.1007/s11063-021-10514-w} (\bibinfo{year}{2021}).

\bibitem{8620826}
\bibinfo{author}{Swief, A.} \& \bibinfo{author}{El-Habrouk, M.}
\newblock \bibinfo{title}{A survey of automotive driving assistance systems
  technologies}.
\newblock In \emph{\bibinfo{booktitle}{2018 International Conference on
  Artificial Intelligence and Data Processing (IDAP)}}, \bibinfo{pages}{1--12},
  \doiprefix\url{10.1109/IDAP.2018.8620826} (\bibinfo{year}{2018}).

\bibitem{rashid_sulaiman_p_p_p_abdul_majeed_musa_ab_nasir_bari_khatun_2020}
\bibinfo{author}{Rashid, M.} \emph{et~al.}
\newblock \bibinfo{journal}{\bibinfo{title}{Current status, challenges, and
  possible solutions of eeg-based brain-computer interface: A comprehensive
  review}}.
\newblock {\emph{\JournalTitle{Frontiers in Neurorobotics}}}
  \textbf{\bibinfo{volume}{14}}, \doiprefix\url{10.3389/fnbot.2020.00025}
  (\bibinfo{year}{2020}).

\bibitem{lawhern_2018}
\bibinfo{author}{Lawhern, V.~J.}, \emph{et~al.}
\newblock \bibinfo{journal}{\bibinfo{title}{EEGNet: a compact convolutional neural 
 network for EEG-based brain–computer interfaces}}.
\newblock {\emph{\JournalTitle{Journal of Neural Engineering}}}
  \textbf{\bibinfo{volume}{15}}, \bibinfo{pages}{056013},
  \doiprefix\url{10.1088/1741-2552/aace8c} (\bibinfo{year}{2018}).

  \bibitem{chuang_lai_ko_kuo_lin_2010}
\bibinfo{author}{Chuang, C.-H.}, \bibinfo{author}{Lai, P.-C.},
  \bibinfo{author}{Ko, L.-W.}, \bibinfo{author}{Kuo, B.-C.} \&
  \bibinfo{author}{Lin, C.-T.}
\newblock \bibinfo{journal}{\bibinfo{title}{Driver's cognitive state
  classification toward brain computer interface via using a generalized and
  supervised technology}}.
\newblock {\emph{\JournalTitle{The 2010 International Joint Conference on
  Neural Networks (IJCNN)}}} \doiprefix\url{10.1109/ijcnn.2010.5596835}
  (\bibinfo{year}{2010}).

\bibitem{lin_ko_chen_chen_lin_2010}
\bibinfo{author}{Lin, F.-C.}, \bibinfo{author}{Ko, L.-W.},
  \bibinfo{author}{Chen, S.-A.}, \bibinfo{author}{Chen, C.-F.} \&
  \bibinfo{author}{Lin, C.-T.}
\newblock \bibinfo{journal}{\bibinfo{title}{Eeg-based cognitive state
  monitoring and predition by using the self-constructing neural fuzzy
  system}}.
\newblock {\emph{\JournalTitle{Proceedings of 2010 IEEE International Symposium
  on Circuits and Systems}}} \doiprefix\url{10.1109/iscas.2010.5536955}
  (\bibinfo{year}{2010}).

\bibitem{zheng_lu_2017}
\bibinfo{author}{Zheng, W.-L.} \& \bibinfo{author}{Lu, B.-L.}
\newblock \bibinfo{journal}{\bibinfo{title}{A multimodal approach to estimating
  vigilance using eeg and forehead eog}}.
\newblock {\emph{\JournalTitle{Journal of Neural Engineering}}}
  \textbf{\bibinfo{volume}{14}}, \bibinfo{pages}{026017},
  \doiprefix\url{10.1088/1741-2552/aa5a98} (\bibinfo{year}{2017}).

\bibitem{hajinoroozi_zhang_huang_2017}
\bibinfo{author}{Hajinoroozi, M.}, \bibinfo{author}{Zhang, J.} \&
  \bibinfo{author}{Huang, Y.}
\newblock \bibinfo{journal}{\bibinfo{title}{Prediction of fatigue-related
  driver performance from eeg data by deep riemannian model}}.
\newblock {\emph{\JournalTitle{2017 39th Annual International Conference of the
  IEEE Engineering in Medicine and Biology Society (EMBC)}}}
  \doiprefix\url{10.1109/embc.2017.8037774} (\bibinfo{year}{2017}).

\bibitem{10.3389/fneng.2012.00013}
\bibinfo{author}{Lew, E.}, \bibinfo{author}{Chavarriaga, R.},
  \bibinfo{author}{Silvoni, S.} \& \bibinfo{author}{Millán, J. d.~R.}
\newblock \bibinfo{journal}{\bibinfo{title}{Detection of self-paced reaching
  movement intention from eeg signals}}.
\newblock {\emph{\JournalTitle{Frontiers in Neuroengineering}}}
  \textbf{\bibinfo{volume}{5}}, \doiprefix\url{10.3389/fneng.2012.00013}
  (\bibinfo{year}{2012}).

\bibitem{garipelli_chavarriaga_del_r_millan_2013}
\bibinfo{author}{Garipelli, G.}, \bibinfo{author}{Chavarriaga, R.} \&
  \bibinfo{author}{del R~Millan, J.}
\newblock \bibinfo{journal}{\bibinfo{title}{Single trial analysis of slow
  cortical potentials: A study on anticipation related potentials}}.
\newblock {\emph{\JournalTitle{Journal of Neural Engineering}}}
  \textbf{\bibinfo{volume}{10}}, \bibinfo{pages}{036014},
  \doiprefix\url{10.1088/1741-2560/10/3/036014} (\bibinfo{year}{2013}).

\bibitem{10.3389/fnins.2014.00222}
\bibinfo{author}{Lew, E. Y.~L.}, \bibinfo{author}{Chavarriaga, R.},
  \bibinfo{author}{Silvoni, S.} \& \bibinfo{author}{Millán, J. d.~R.}
\newblock \bibinfo{journal}{\bibinfo{title}{Single trial prediction of
  self-paced reaching directions from eeg signals}}.
\newblock {\emph{\JournalTitle{Frontiers in Neuroscience}}}
  \textbf{\bibinfo{volume}{8}}, \doiprefix\url{10.3389/fnins.2014.00222}
  (\bibinfo{year}{2014}).

\bibitem{haufe_treder_gugler_sagebaum_curio_blankertz_2011}
\bibinfo{author}{Haufe, S.} \emph{et~al.}
\newblock \bibinfo{journal}{\bibinfo{title}{Eeg potentials predict upcoming
  emergency brakings during simulated driving}}.
\newblock {\emph{\JournalTitle{Journal of Neural Engineering}}}
  \textbf{\bibinfo{volume}{8}}, \bibinfo{pages}{056001},
  \doiprefix\url{10.1088/1741-2560/8/5/056001} (\bibinfo{year}{2011}).

\bibitem{khaliliardali_chavarriaga_gheorghe_millan_2015}
\bibinfo{author}{Khaliliardali, Z.}, \bibinfo{author}{Chavarriaga, R.},
  \bibinfo{author}{Gheorghe, L.~A.} \& \bibinfo{author}{Millan, J.~d.}
\newblock \bibinfo{journal}{\bibinfo{title}{Action prediction based on
  anticipatory brain potentials during simulated driving}}.
\newblock {\emph{\JournalTitle{Journal of Neural Engineering}}}
  \textbf{\bibinfo{volume}{12}}, \bibinfo{pages}{066006},
  \doiprefix\url{10.1088/1741-2560/12/6/066006} (\bibinfo{year}{2015}).

\bibitem{9669864}
\bibinfo{author}{Honzík, V.} \& \bibinfo{author}{Mouček, R.}
\newblock \bibinfo{title}{Spiking neural networks for classification of
  brain-computer interface and image data}.
\newblock In \emph{\bibinfo{booktitle}{2021 IEEE International Conference on
  Bioinformatics and Biomedicine (BIBM)}}, \bibinfo{pages}{3624--3629},
  \doiprefix\url{10.1109/BIBM52615.2021.9669864} (\bibinfo{year}{2021}).

\bibitem{9629621}
\bibinfo{author}{Pals, M.} \emph{et~al.}
\newblock \bibinfo{title}{Demonstrating the viability of mapping deep learning
  based eeg decoders to spiking networks on low-powered neuromorphic chips}.
\newblock In \emph{\bibinfo{booktitle}{2021 43rd Annual International
  Conference of the IEEE Engineering in Medicine Biology Society (EMBC)}},
  \bibinfo{pages}{6102--6105}, \doiprefix\url{10.1109/EMBC46164.2021.9629621}
  (\bibinfo{year}{2021}).

\bibitem{9024211}
\bibinfo{author}{Luo, Y.} \emph{et~al.}
\newblock \bibinfo{journal}{\bibinfo{title}{Eeg-based emotion classification
  using spiking neural networks}}.
\newblock {\emph{\JournalTitle{IEEE Access}}} \textbf{\bibinfo{volume}{8}},
  \bibinfo{pages}{46007--46016}, \doiprefix\url{10.1109/ACCESS.2020.2978163}
  (\bibinfo{year}{2020}).

\bibitem{singanamalla_lin_2021}
\bibinfo{author}{Singanamalla, S.~K.} \& \bibinfo{author}{Lin, C.-T.}
\newblock \bibinfo{journal}{\bibinfo{title}{Spiking neural network for
  augmenting electroencephalographic data for brain computer interfaces}}.
\newblock {\emph{\JournalTitle{Frontiers in Neuroscience}}}
  \textbf{\bibinfo{volume}{15}}, \doiprefix\url{10.3389/fnins.2021.651762}
  (\bibinfo{year}{2021}).

\bibitem{kumarasinghe_kasabov_taylor_2021}
\bibinfo{author}{Kumarasinghe, K.}, \bibinfo{author}{Kasabov, N.} \&
  \bibinfo{author}{Taylor, D.}
\newblock \bibinfo{journal}{\bibinfo{title}{Brain-inspired spiking neural
  networks for decoding and understanding muscle activity and kinematics from
  electroencephalography signals during hand movements}}.
\newblock {\emph{\JournalTitle{Scientific Reports}}}
  \textbf{\bibinfo{volume}{11}}, \doiprefix\url{10.1038/s41598-021-81805-4}
  (\bibinfo{year}{2021}).

\bibitem{yan_zhou_wong_2022}
\bibinfo{author}{Yan, Z.}, \bibinfo{author}{Zhou, J.} \& \bibinfo{author}{Wong,
  W.-F.}
\newblock \bibinfo{journal}{\bibinfo{title}{Eeg classification with spiking
  neural network: Smaller, better, more energy efficient}}.
\newblock {\emph{\JournalTitle{Smart Health}}} \textbf{\bibinfo{volume}{24}},
  \bibinfo{pages}{100261}, \doiprefix\url{10.1016/j.smhl.2021.100261}
  (\bibinfo{year}{2022}).

\bibitem{shah_wang_doborjeh_doborjeh_kasabov_2019}
\bibinfo{author}{Shah, D.}, \bibinfo{author}{Wang, G.~Y.},
  \bibinfo{author}{Doborjeh, M.}, \bibinfo{author}{Doborjeh, Z.} \&
  \bibinfo{author}{Kasabov, N.}
\newblock \bibinfo{journal}{\bibinfo{title}{Deep learning of eeg data in the
  neucube brain-inspired spiking neural network architecture for a better
  understanding of depression}}.
\newblock {\emph{\JournalTitle{Neural Information Processing}}}
  \bibinfo{pages}{195--206}, \doiprefix\url{10.1007/978-3-030-36718-3_17}
  (\bibinfo{year}{2019}).

\bibitem{9721187}
\bibinfo{author}{Chen, X.} \emph{et~al.}
\newblock \bibinfo{journal}{\bibinfo{title}{Toward open-world
  electroencephalogram decoding via deep learning: A comprehensive survey}}.
\newblock {\emph{\JournalTitle{IEEE Signal Processing Magazine}}}
  \textbf{\bibinfo{volume}{39}}, \bibinfo{pages}{117--134},
  \doiprefix\url{10.1109/MSP.2021.3134629} (\bibinfo{year}{2022}).

\bibitem{merolla_arthur_alvarez_icaza_cassidy_sawada_akopyan_jackson_imam_guo_nakamura_etal_2014}
\bibinfo{author}{Merolla, P.~A.} \emph{et~al.}
\newblock \bibinfo{journal}{\bibinfo{title}{A million spiking-neuron integrated
  circuit with a scalable communication network and interface}}.
\newblock {\emph{\JournalTitle{Science}}} \textbf{\bibinfo{volume}{345}},
  \bibinfo{pages}{668--673}, \doiprefix\url{10.1126/science.1254642}
  (\bibinfo{year}{2014}).

\bibitem{jo_chang_ebong_bhadviya_mazumder_lu_2010}
\bibinfo{author}{Jo, S.~H.} \emph{et~al.}
\newblock \bibinfo{journal}{\bibinfo{title}{Nanoscale memristor device as
  synapse in neuromorphic systems}}.
\newblock {\emph{\JournalTitle{Nano Letters}}} \textbf{\bibinfo{volume}{10}},
  \bibinfo{pages}{1297--1301}, \doiprefix\url{10.1021/nl904092h}
  (\bibinfo{year}{2010}).

\bibitem{DBLP:journals/corr/abs-1906-08853}
\bibinfo{author}{Gopalakrishnan, R.}, \bibinfo{author}{Chua, Y.} \&
  \bibinfo{author}{Kumar, A. J.~S.}
\newblock \bibinfo{journal}{\bibinfo{title}{Hardware-friendly neural network
  architecture for neuromorphic computing}}.
\newblock {\emph{\JournalTitle{CoRR}}}
  \textbf{\bibinfo{volume}{abs/1906.08853}} (\bibinfo{year}{2019}).
\newblock \eprint{1906.08853}.

\bibitem{ivanov_2022}
\bibinfo{author}{Ivanov, D.}, \bibinfo{author}{Chezhegov, A.}, 
  \bibinfo{author}{Kiselev, M.}, \bibinfo{author}{Grunin, A.} \&
  \bibinfo{author}{Larionov, D.}
\newblock \bibinfo{journal}{\bibinfo{title}{Neuromorphic artificial intelligence
  systems}}.
\newblock {\emph{\JournalTitle{Frontiers in Neuroscience}}}
  \textbf{\bibinfo{volume}{16}}, \doiprefix\url{10.3389/fnins.2022.959626}
  (\bibinfo{year}{2022}).

\bibitem{scherr_stock1_maass_2020}

\bibinfo{author}{Scherr, F., Stöckl, C. \& Maass, W.}
\newblock \bibinfo{title}{One-shot learning with spiking neural networks}.
\newblock \bibinfo{journal}{\emph{\JournalTitle{bioRxiv}}}. \doiprefix\url{10.1101/2020.06.17.156513} 
(\bibinfo{year}{2020}).

\bibitem{demir_koike-akino_wang_haruna_erdogmus_2021}
\bibinfo{author}{Demir, A.}, \bibinfo{author}{Koike-Akino, T.},
  \bibinfo{author}{Wang, Y.}, \bibinfo{author}{Haruna, M.} \&
  \bibinfo{author}{Erdogmus, D.}
\newblock \bibinfo{journal}{\bibinfo{title}{Eeg-gnn: Graph neural networks for
  classification of electroencephalogram (eeg) signals}}.
\newblock {\emph{\JournalTitle{2021 43rd Annual International Conference of the
  IEEE Engineering in Medicine \& Biology Society (EMBC)}}}
  \doiprefix\url{10.1109/embc46164.2021.9630194} (\bibinfo{year}{2021}).

\bibitem{https://doi.org/10.48550/arxiv.2007.13484}
\bibinfo{author}{Jia, S.}, \bibinfo{author}{Hou, Y.}, \bibinfo{author}{Shi, Y.}
  \& \bibinfo{author}{Li, Y.}
\newblock \bibinfo{title}{Attention-based graph resnet for motor intent
  detection from raw eeg signals}, \doiprefix\url{10.48550/ARXIV.2007.13484}
  (\bibinfo{year}{2020}).

\bibitem{https://doi.org/10.48550/arxiv.2006.08924}
\bibinfo{author}{Hou, Y.} \emph{et~al.}
\newblock \bibinfo{title}{Gcns-net: A graph convolutional neural network
  approach for decoding time-resolved eeg motor imagery signals},
  \doiprefix\url{10.48550/ARXIV.2006.08924} (\bibinfo{year}{2020}).

\bibitem{https://doi.org/10.48550/arxiv.1907.07835}
\bibinfo{author}{Zhong, P.}, \bibinfo{author}{Wang, D.} \&
  \bibinfo{author}{Miao, C.}
\newblock \bibinfo{title}{Eeg-based emotion recognition using regularized graph
  neural networks}, \doiprefix\url{10.48550/ARXIV.1907.07835}
  (\bibinfo{year}{2019}).

\bibitem{teng_bi_liu_2018}
\bibinfo{author}{Teng, T.}, \bibinfo{author}{Bi, L.} \& \bibinfo{author}{Liu,
  Y.}
\newblock \bibinfo{journal}{\bibinfo{title}{Eeg-based detection of driver
  emergency braking intention for brain-controlled vehicles}}.
\newblock {\emph{\JournalTitle{IEEE Transactions on Intelligent Transportation
  Systems}}} \textbf{\bibinfo{volume}{19}}, \bibinfo{pages}{1766--1773},
  \doiprefix\url{10.1109/tits.2017.2740427} (\bibinfo{year}{2018}).

\bibitem{kim_kim_haufe_lee_2014}
\bibinfo{author}{Kim, I.-H.}, \bibinfo{author}{Kim, J.-W.},
  \bibinfo{author}{Haufe, S.} \& \bibinfo{author}{Lee, S.-W.}
\newblock \bibinfo{journal}{\bibinfo{title}{Detection of braking intention in
  diverse situations during simulated driving based on eeg feature
  combination}}.
\newblock {\emph{\JournalTitle{Journal of Neural Engineering}}}
  \textbf{\bibinfo{volume}{12}}, \bibinfo{pages}{016001},
  \doiprefix\url{10.1088/1741-2560/12/1/016001} (\bibinfo{year}{2014}).

\bibitem{https://www.biorxiv.org/content/10.1101/443390v1}
\bibinfo{author}{Khaliliardali, Z.} \emph{et~al.}
\newblock \bibinfo{title}{Real-time detection of driver's movement intention in
  response to traffic lights}, \doiprefix\url{10.1101/443390}
  (\bibinfo{year}{2019}).

\bibitem{hernandez_mozos_ferrandez_antelis_2018}
\bibinfo{author}{Hernandez, L.~G.}, \bibinfo{author}{Mozos, O.~M.},
  \bibinfo{author}{Ferrandez, J.~M.} \& \bibinfo{author}{Antelis, J.~M.}
\newblock \bibinfo{journal}{\bibinfo{title}{Eeg-based detection of braking
  intention under different car driving conditions}}.
\newblock {\emph{\JournalTitle{Frontiers in Neuroinformatics}}}
  \textbf{\bibinfo{volume}{12}}, \doiprefix\url{10.3389/fninf.2018.00029}
  (\bibinfo{year}{2018}).

\bibitem{nguyen_chung_2019}
\bibinfo{author}{Nguyen, T.-H.} \& \bibinfo{author}{Chung, W.-Y.}
\newblock \bibinfo{journal}{\bibinfo{title}{Detection of driver braking
  intention using eeg signals during simulated driving}}.
\newblock {\emph{\JournalTitle{Sensors}}} \textbf{\bibinfo{volume}{19}},
  \bibinfo{pages}{2863}, \doiprefix\url{10.3390/s19132863}
  (\bibinfo{year}{2019}).

\bibitem{lee_kim_lee_2017}
\bibinfo{author}{Lee, S.-M.}, \bibinfo{author}{Kim, J.-W.} \&
  \bibinfo{author}{Lee, S.-W.}
\newblock \bibinfo{journal}{\bibinfo{title}{Detecting driver's braking
  intention using recurrent convolutional neural networks based eeg analysis}}.
\newblock {\emph{\JournalTitle{2017 4th IAPR Asian Conference on Pattern
  Recognition (ACPR)}}} \doiprefix\url{10.1109/acpr.2017.86}
  (\bibinfo{year}{2017}).

\bibitem{s141018172}
Planelles, D., Hortal, E., Costa, Á., Úbeda, A., Iáez, E. \& Azorín, J. Evaluating Classifiers to Detect Arm Movement Intention from EEG Signals. {\em Sensors}. \textbf{14}, 18172-18186 (2014), https://www.mdpi.com/1424-8220/14/10/18172

\bibitem{7930496}
Chamanzar, A., Shabany, M., Malekmohammadi, A. \& Mohammadinejad, S. Efficient Hardware Implementation of Real-Time Low-Power Movement Intention Detector System Using FFT and Adaptive Wavelet Transform. {\em IEEE Transactions On Biomedical Circuits And Systems}. \textbf{11}, 585-596 (2017)

\bibitem{8616177}
Song, M., Oh, S., Jeong, H., Kim, J. \& Kim, J. A Novel Movement Intention Detection Method for Neurorehabilitation Brain-Computer Interface System. {\em 2018 IEEE International Conference On Systems, Man, And Cybernetics (SMC)}. pp. 1016-1021 (2018)

\bibitem{Nguyen16397}
Nguyen, V., Breakspear, M. \& Cunnington, R. Reciprocal Interactions of the SMA and Cingulate Cortex Sustain Premovement Activity for Voluntary Actions. {\em Journal Of Neuroscience}. \textbf{34}, 16397-16407 (2014), https://www.jneurosci.org/content/34/49/16397

\bibitem{MIRZABAGHERIAN2023107159}

Mirzabagherian, H., Menhaj, M., Suratgar, A., Talebi, N., Abbasi Sardari, M. \& Sajedin, A. Temporal-spatial convolutional residual network for decoding attempted movement related EEG signals of subjects with spinal cord injury. {\em Computers In Biology And Medicine}. \textbf{164} pp. 107159 (2023), https://www.sciencedirect.com/science/article/pii/S0010482523006248

\bibitem{Gatti492660}
Gatti, R., Atum, Y., Schiaffino, L., Jochumsen, M. \& Manresa, J. Prediction of Hand Movement Speed and Force from Single-trial EEG with Convolutional Neural Networks. {\em BioRxiv}. (2019), https://www.biorxiv.org/content/early/2019/11/07/492660

\bibitem{PPR:PPR485383}
Di Russo, F. \& Mussini, E. Reduction of anticipatory brain activity in anxious people and regulatory effect of response-related feedback. (PsyArXiv,2022), https://doi.org/10.31234/osf.io/ekh45

\bibitem{grattarola_alippi_2021}
\bibinfo{author}{Grattarola, D.} \& \bibinfo{author}{Alippi, C.}
\newblock \bibinfo{journal}{\bibinfo{title}{Graph neural networks in tensorflow
  and keras with spektral [application notes]}}.
\newblock {\emph{\JournalTitle{IEEE Computational Intelligence Magazine}}}
  \textbf{\bibinfo{volume}{16}}, \bibinfo{pages}{99--106},
  \doiprefix\url{10.1109/mci.2020.3039072} (\bibinfo{year}{2021}).

\bibitem{NEURIPS2019_9015}
\bibinfo{author}{Paszke, A.} \emph{et~al.}
\newblock \bibinfo{title}{Pytorch: An imperative style, high-performance deep
  learning library}.
\newblock In \bibinfo{editor}{Wallach, H.} \emph{et~al.} (eds.)
  \emph{\bibinfo{booktitle}{Advances in Neural Information Processing Systems
  32}}, \bibinfo{pages}{8024--8035} (\bibinfo{publisher}{Curran Associates,
  Inc.}, \bibinfo{year}{2019}).

\bibitem{tensorflow2015-whitepaper}
\bibinfo{author}{Abadi, M.} \emph{et~al.}
\newblock \bibinfo{title}{TensorFlow: Large-scale machine learning on heterogeneous systems}.
\newblock \bibinfo{howpublished}{https://www.tensorflow.org/}. (\bibinfo{year}{2015}).


\bibitem{eshraghian2021training}
\bibinfo{author}{Eshraghian, J.~K.} \emph{et~al.}
\newblock \bibinfo{journal}{\bibinfo{title}{Training spiking neural networks
  using lessons from deep learning}}.
\newblock {\emph{\JournalTitle{arXiv preprint arXiv:2109.12894}}}
  (\bibinfo{year}{2021}).

\bibitem{wilson_2015}
\bibinfo{author}{Wilson, R.~J.}
\newblock \emph{\bibinfo{title}{Introduction to graph theory}}
  (\bibinfo{publisher}{Prentice Hall}, \bibinfo{year}{2015}).

\bibitem{wu_pan_chen_long_zhang_yu_2021}
\bibinfo{author}{Wu, Z.} \emph{et~al.}
\newblock \bibinfo{journal}{\bibinfo{title}{A comprehensive survey on graph
  neural networks}}.
\newblock {\emph{\JournalTitle{IEEE Transactions on Neural Networks and
  Learning Systems}}} \textbf{\bibinfo{volume}{32}}, \bibinfo{pages}{4--24},
  \doiprefix\url{10.1109/tnnls.2020.2978386} (\bibinfo{year}{2021}).

\bibitem{https://doi.org/10.48550/arxiv.1609.02907}
\bibinfo{author}{Kipf, T.~N.} \& \bibinfo{author}{Welling, M.}
\newblock \bibinfo{title}{Semi-supervised classification with graph
  convolutional networks}, \doiprefix\url{10.48550/ARXIV.1609.02907}
  (\bibinfo{year}{2016}).

\bibitem{https://doi.org/10.48550/arxiv.1810.00826}
\bibinfo{author}{Xu, K.}, \bibinfo{author}{Hu, W.}, \bibinfo{author}{Leskovec,
  J.} \& \bibinfo{author}{Jegelka, S.}
\newblock \bibinfo{title}{How powerful are graph neural networks?},
  \doiprefix\url{10.48550/ARXIV.1810.00826} (\bibinfo{year}{2018}).

\bibitem{Neuroelectrics}
\bibinfo{title}{Neuroelectrics}.
\newblock \bibinfo{howpublished}{https://www.neuroelectrics.com}.
\newblock \bibinfo{note}{Accessed: 18-May-2022}.

\bibitem{delorme_makeig_2004}
\bibinfo{author}{Delorme, A.} \& \bibinfo{author}{Makeig, S.}
\newblock \bibinfo{journal}{\bibinfo{title}{Eeglab: An open source toolbox for
  analysis of single-trial eeg dynamics including independent component
  analysis}}.
\newblock {\emph{\JournalTitle{Journal of Neuroscience Methods}}}
  \textbf{\bibinfo{volume}{134}}, \bibinfo{pages}{9--21},
  \doiprefix\url{10.1016/j.jneumeth.2003.10.009} (\bibinfo{year}{2004}).

\bibitem{mattan_s_ben_shachar_2022_5948294}
\bibinfo{author}{Ben-Shachar, M.~S.}
\newblock \bibinfo{title}{mattansb/tbt: Channel you inner error},
  \doiprefix\url{10.5281/zenodo.5948294} (\bibinfo{year}{2022}).

\bibitem{bagdasarov_roberts_brechet_brunet_michel_gaffrey_2022}
\bibinfo{author}{Bagdasarov, A.} \emph{et~al.}
\newblock \bibinfo{title}{Spatiotemporal dynamics of eeg microstates in four-
  to eight-year-old children: Age- and sex-related effects},
  \doiprefix\url{10.31234/osf.io/x35uf} (\bibinfo{year}{2022}).

\end{thebibliography}

\section*{Author contributions}

 N.L, V.S.S.N. and K.K. conceptualized and designed the method, performed analysis and prepared the manuscript. N.L. collected and processed the EEG data, and developed the software code to generate the results.

\section*{Competing interests} 

The authors declare that they have no competing interests.

\newpage

\section*{Legend}

Table 1. Classification performance with floating-point EEG input data (best performance in each classification measure highlighted in bold font)
\newline \newline
Table 2. Five-channel ablation study with floating-point EEG input data (best performance in each classification measure highlighted in bold font)
\newline \newline
Table 3. Classification performance with delta modulated spike train input data (best performance in each classification measure highlighted in bold font
\newline \newline
Figure 1. Pre-Processed EEG Signals. (a) Channel potentials with associated countdown and "Stop" command markers and scale. (b) Cz grand average signal with scalp maps representing the grand average at the midpoint between two neighboring markers and color bar on the right displaying the potential in $\mu V$.
\newline \newline
Figure 2. Schematic of neural network architectures. (a) CSNN. (b) CNN.
\newline \newline
Figure 3. Schematic diagram of EEGNet.
\newline \newline
Figure 4. Schematic diagrams of GNNs. (a) GCNConv. (b) GCSConv. (c) GINConv.
\newline \newline
Figure 5. Experimental design illustration of a trial. Photo by Micheal Pierce/Missouri S\&T
\newline \newline
Figure 6. Data processing pipeline.

\end{document}